\documentclass{article}
\usepackage{graphicx}
\usepackage{helvet}
\usepackage{courier}
\usepackage{latexsym}

\begin{document}

\newtheorem{theorem}{Theorem}
\newtheorem{obs}{Observation}
\newtheorem{corollary}{Corollary}
\newtheorem{lemma}{Lemma}
\newtheorem{definition}{Definition}
\newcommand{\proof}{\noindent {\em Proof.\ \ }}
\newcommand{\qed}{\hfill \mbox{$\Box$}\medskip}
\newcommand{\mymod}{\, \mbox{mod} \, }
\newcommand{\mydiv}{\, \mbox{div} \, }
\def\mymax{\mbox{$max$}}
\def\mymin{\mbox{$min$}}
\newcommand{\nvalue}{{\tt NValue}}
\newcommand{\common}{{\tt Common}}
\newcommand{\disjoint}{{\tt Disjoint}}
\newcommand{\atmost}{{\tt AtMost}}
\newcommand{\atleast}{{\tt AtLeast}}
\newcommand{\atmostone}{{\tt AtMost1}}
\newcommand{\sprod}{{\tt ScalarProduct}}
\newcommand{\alldiff}{{\tt AllDifferent}}
\newcommand{\among}{{\tt Among}}
\newcommand{\gcc}{{\tt Gcc}}

\newcommand{\card}{{\tt Card}}
\newcommand{\cardpath}{{\tt Cardpath}}

\title{The Complexity of Reasoning with Global Constraints}

\author{
Christian Bessiere\\
LIRMM, CNRS/U. Montpellier \\
Montpellier, France \\
bessiere@lirmm.fr
\and
Emmanuel Hebrard\\
4C and UCC\\ 
Cork, Ireland\\
e.hebrard@4c.ucc.ie
\and
Brahim Hnich\\
Izmir University of Economics\\
Izmir, Turkey\\
brahim.hnich@ieu.edu.tr
\and
Toby Walsh\\
NICTA and UNSW\\
Sydney, Australia\\
tw@cse.unsw.edu.au
}

\maketitle
\begin{abstract}
Constraint propagation is one of the techniques central
to the success of constraint programming. To reduce search,
fast algorithms associated with each constraint 
prune the domains of variables. With global (or non-binary) constraints, 
the cost of such propagation may be much greater than the quadratic
cost for binary constraints. We therefore
study the computational complexity of reasoning with
global constraints. We first characterise a number of important
questions related to constraint propagation. We show that 
such questions are intractable in general, 
and identify dependencies between the tractability and
intractability of the different questions. % for finite domain variables. 
We then demonstrate how the tools of
computational complexity can be used in the design and analysis of
specific global constraints.  In particular, we illustrate how
computational complexity can be used to determine when a lesser level
of local consistency should be enforced, when constraints 
can be safely generalized, when decomposing constraints
will reduce the amount of pruning, and 
when combining constraints is tractable.  
%We then show that with set or multiset variables, constraint propagation can 
%be intractable even on bounded arity constraints. 
%Finally, we show that the same
%tools can be used to  study symmetry breaking,
%meta-constraints like the cardinality constraint. 
\end{abstract}

\section{Introduction}
Constraint programming is a very successful technology for solving many
kinds of combinatorial problems arising in industrial applications,
such as scheduling, resource allocation, vehicle routing, and
product configuration \cite{wallace96}. 
One of its key features is \emph{constraint propagation} where
values in the domains of variables are removed which
will lead to a constraint violation. 
Constraint propagation can prune large parts of the
search space, and is vital for solving combinatorially 
challenging problems.  
The notion of local consistency provides a formal way to characterise
the amount of work done by constraint propagation. 
The most common level of local
consistency, called generalised arc consistency (GAC), specifies that
all values inconsistent with a constraint are pruned.  

Constraint propagation on binary (or bounded arity)
constraints is polynomial. However, constraint toolkits
support an increasing number of global (or non-binary) constraints since
such constraints are central to the success
of constraint programming. See, for example,
\cite{regin1,regin2,Bessiere-Regin97,regin4,beldiceanu2,fhkmwcp2002}. 
Global constraints specify patterns that occur in many problems, and
use  constraint propagation
algorithms  that exploit their precise semantics. They  permit users
to model  problems compactly
and solvers to prune the search space effectively. 
%%%christian
They often allow
efficient propagation. 
For instance, we often have sets of variables which must
take different values (e.g. activities in a scheduling problem
requiring the same resource must all be assigned different
times). Most constraint solvers therefore provide a global
'$\alldiff$' constraint which is propagated efficiently
and effectively \cite{KnuRag92,regin1}.
In many problems, the arity of such global constraints
can grow with the problem size. For example, 
in the Golomb ruler problem (prob006 in CSPLib), the
size of the $\alldiff$ constraint grows quadratically with the 
number of ticks on the ruler.
Similarly, in the balanced incomplete
block design (prob028 in CSPLib), the size of the intersection
constraint between rows grows linearly with the number of blocks. 
Such global constraints
may therefore exhibit complexities far beyond the quadratic
cost for propagating binary constraints. 

What then are the limits of reasoning
with global constraints? In this paper, we
show how the basic tools of computational complexity
can be used to uncover many of the basic limits. 
We characterise the different reasoning
tasks related to constraint propagation. For example,
``is this value consistent with this constraint?'' or 
%``what are the maximal generalised arc-consistent domains?''
``do there exist values consistent with this constraint?''. 
We identify dependencies between the
tractability and intractability of these different
questions. We
show that all of them are
intractable in general. We therefore need to focus
on specific constraints like the $\alldiff$ constraint
which are tractable. We then show how these same tools
of computational complexity can be used to
analyse specific global constraints proposed in
the past like the number of values constraint \cite{pachet1},
as well as to help design new global constraints. 
%We also show that when dealing with set
%variables, constraint propagation can become intractable even on
%bounded arity constraints.  
%Afterwards, 
% We use these results
% to study a range of existing and new global 
% constraints on integer %, set and multiset
% variables. We either show that the global constraints are tractable (by
% giving a polynomial algorithm or referencing one in the
% literature) or prove that they are NP-hard to propagate. 
% In the later case, we expect that any
% decomposition will hinder propagation (unless P=NP). 

Computational complexity provides
a methodology to decide when a lesser level of
propagation should be enforced or when decomposing a constraint
hinders propagation. It also tells us whether a new global constraint
designed as a combination of elementary constraints or as a
generalisation of an existing tractable constraint will itself be
tractable. 
%allow polynomial  algorithms for generalised arc consistency. 
%Finally, those tools can be applied to the analysis of other 
%notions essential in constraint programming, such as symmetry
%detection and nogood learning. 

The rest of the paper is organised as follows. Section \ref{sec:back}
presents the technical background necessary to read the subsequent
sections. Section \ref{sec:gac} contains a theoretical study of
generalised arc consistency, the central notion of local consistency
used when speaking of constraint propagation. In Section
\ref{sec:classify}, we show how the tools of computational complexity
can be used to analyse different types of global constraints. An extension to
meta-constraints (constraints that must be satisfied a given number of
times) is presented in Section \ref{sec:meta}. 
%Sections \ref{sec:sym}
%and \ref{sec:nogoods} apply the same tools to symmetry breaking and
%nogood learning issues. 
Finally Section \ref{sec:related} discusses
related work and Section \ref{sec:conc} concludes the paper.

\section{Theoretical  Background}\label{sec:back}
A {\em constraint satisfaction problem} (\textsc{CSP}) consists of
a set of variables, each with a finite
domain of values, and a set of
constraints that specify allowed combinations of values for subsets of
variables. We will denote variables with upper case letters and
values with lower case. We will assume that the domain of a variable
is given extensionally, but that a constraint $C$ is given intensionally by
a function of the form  $f_C: D(X_1) \times \ldots \times D(X_n) \mapsto
\{true,false\}$ where $D(X_i)$ are the domains of the
variables in the scope $var(C)=(X_1,\ldots,X_n)$ of the constraint $C$.
We say that $D$ is a domain on $var(C)$.
We cannot permit an arbitrary sort of function. 
For example, suppose $f_C(3,1,5,2,3,1,\ldots)$
returns $true$ iff the \textsc{1in3-3SAT} problem,
${x_3} \vee {x_1} \vee {x_5}, x_2 \vee x_3 \vee x_1, \ldots$ is satisfiable.
Testing if an assignment satisfies this non-binary constraint
is then NP-complete, and finding a satisfying assignment
is PSPACE-complete. As a second example,
suppose domains are integers of size $m$ and $f_C(X_1,X_2,X_3,\ldots)$ is the
function that halts iff $X_1+X_2*m+X_3*m^2+\ldots$ is the
G\"odel number of a halting Turing machine.
Even testing if an assignment satisfies such a constraint
is undecidable. We therefore insist that
$f_C$ is computable in polynomial time.

Constraint toolkits usually contain a library of predefined
\emph{constraint types} with a particular semantics that can be
applied to sets of 
variables with varying arities and  domains. 
A constraint is only an instance of a
constraint type on given variables and domains. 
For instance, $\alldiff$ is a constraint
type. $\alldiff(X_1,..,X_3)$ with
$D(X_1)=D(X_2)=\{1,2\},D(X_3)=\{1,2,3\}$ is an instance of constraint
of the type $\alldiff$. When there is no ambiguity, we will use the
terms 'constraint' or 'constraint type' indifferently.  

A solution to a \textsc{CSP}
is an assignment of values to the
variables satisfying the constraints. To find such
solutions, we %constraint solvers
often use tree search algorithms that construct
partial assignments and enforce a local consistency 
%like generalised arc consistency 
to prune the search space. 
Enforcing a local consistency is traditionally called
\emph{constraint propagation}.
One of the most commonly used local consistencies
is generalised arc consistency. A constraint $C$ is
{\em  generalised arc consistent} (GAC) iff, when a variable in the
scope of $C$ is assigned any value in its domain, there exists an assignment to the
other variables in $C$ such that $C$ is satisfied \cite{mohr1}.
This satisfying assignment is called {\em support} for the value.
On binary constraints (those involving just two
variables), generalised arc consistency is
called arc consistency (AC). 

Since this paper makes significant use of computational complexity theory,
we very briefly recall the basic tools for showing
intractability. 
P is the class of decision problems that can be solved by a deterministic
Turing machine in polynomial time, and
NP is the class of decision problems that can be solved by a
non-deterministic 
Turing machine in polynomial time. 
As in \cite{garjon79}, a \emph{transformation} from a
decision problem $Q_1\in$ NP to a decision problem $Q_2\in$ NP is a
function $\varphi$ that polynomially  rewrites
any input $x$ of $Q_1$ into an input $\varphi(x)$ of $Q_2$ such that
$Q_2(\varphi(x))$ answers ``yes'' if and only if $Q_1(x)$ answers
``yes''. If $Q_1$ is NP-complete, this transformation into $Q_2$ permits us
to deduce that $Q_2$ is also NP-complete. 
A \emph{reduction} from a problem $Q_1$ to a problem $Q_2$ (not
necessarily decision problems)  is a program that solves $Q_1$ in
polynomial time under the condition that the program can call
an \emph{oracle} that solves $Q_2$ at most a constant number
of times. If $Q_1$ is NP-complete, this reduction
to $Q_2$ permits to deduce that $Q_2$ is NP-hard. 
%Thus, an NP-hard problem  cannot be solved in polynomial time, unless P=NP. 
Any NP-complete problem is thus NP-hard. 
By transitivity of the
reductions, if $Q_1$ is an NP-hard problem, its reduction
to $Q_2$ permits us to deduce that $Q_2$ is NP-hard. We will
use \emph{intractability} as a 
general term to denote any NP-hard problem, i.e., those 
which cannot be solved in polynomial time unless P=NP. In the
following, we assume P$\neq$ NP. 
A problem in coNP is simply a problem in NP with the answers ``yes''
and ``no'' reversed. 
%coNP is the class of problems taken from NP where  the answers ``yes''
%and ``no'' have been reversed. 
For instance, \textsc{3Sat},  the problem of
deciding if %there exists an interpretation satisfying 
a set of ternary clauses is satisfiable, is in NP. Hence, \textsc{un3Sat}, the
problem of deciding if   a set of ternary clauses is unsatisfiable, is
in coNP. 
The $D^P$ complexity class  contains problems which are the
conjunction of a problem in 
NP and one in coNP  \cite{dpcomplete}.
%modifcation by brahim
% modification of brahim's modification by toby: 
%   we pretty much repeat this in the proof!
%
%More precisely, a language $L$ is in the class $D^P$ iff there are two
%languages $L_1 \in NP$ and $L_2 \in coNP$ such that $L= L_1 \cap
%L_2$. %Note that $D^P$ is \emph{not} $NP \cap coNP$. 
% Given a
% distance matrix and  an integer $B$, {\sc Exact TSP} decides
% if the shortest tour is \emph{equal} to $B$.
% By comparison, \textsc{TSP}
% decides if there is a tour of length $B$ \emph{or less},
% which is NP-complete. 
A problem $Q$ is in $D^P$ if there exist a NP problem $Q_1$ and a coNP
one $Q_2$ such that $Q$ answers ``yes'' iff $Q_1$ and $Q_2$
answer  ``yes''.
If $Q_1$ is NP-complete and $Q_2$ is coNP-complete,
then $Q$ is $D^P$-complete.
The class $D^P$ is also known as the second level of the Boolean hierarchy,
BH$\mbox{}_2$. A typical example of a $D^P$-complete decision
problem is the {\sc Exact Traveling Salesperson Problem} where
we ask if $k$ is the length of the {\em shortest} tour.

\section{Complexity of Generalised Arc Consistency}
\label{sec:gac}

\newcommand{\gacsup}{\textsc{GACSupport}}
\newcommand{\gacisit}{\textsc{IsItGAC}}
\newcommand{\gacwo}{\textsc{NoGACWipeOut}}
\newcommand{\gacmax}{\textsc{maxGAC}}
\newcommand{\gacdom}{\textsc{GACDomain}}

There are different questions that may arise when we consider
enforcing generalised
arc consistency. We can ask whether  a value belongs to a
consistent tuple or whether a constraint is generalised arc consistent. Some of the
questions are more of an academic nature while others are at the heart of
propagation algorithms. In this section, we formally characterise five
questions related to GAC. We study the complexity of GAC reasoning on
global constraints by showing intractability of two of these five
questions. Finally,  we show some dependencies between the
intractability of the questions, from which we conclude  
that all five questions are intractable in general. 

\subsection{Questions related to GAC}
\label{sec:questions}

We characterise five different questions related to reasoning 
about generalized arc consistency. These questions  can be adapted to any other
%'domain filtering' 
local consistency as long as  it rules out values in domains (e.g.,
bounds consistency, singleton arc consistency, etc.) and not
non-unary tuples of values (e.g., path consistency,
relational-$k$-consistency, etc.)  

In the following, \textsc{Problem}($\cal C$) represents the class of questions
defined by \textsc{Problem} on constraints of the
type $\cal C$. \textsc{Problem}($\cal
C$) will %sometimes 
be written \textsc{Problem} when it is not confusing or when there is
no restriction to a particular type of constraints. 
Note also that we use  the notation \textsc{Problem}[data] to
refer to the instance of \textsc{Problem}($\cal C$) with the
input 'data'.  

The first question we consider is at the core of all  generic
arc consistency algorithms. This is the question which is
generally asked for all values one by one. 
\begin{quote}
\gacsup(${\cal C}$)\\
\textbf{Instance.} A constraint $C$ of type $\cal C$, 
  a domain $D$ on $var(C)$, and a
  value $v$ for variable $X$ in $var(C)$\\
\textbf{Question.} Does value $v$ for $X$ have a support on $C$ in $D$?   
\end{quote}

The second question has both practical and theoretical
importance. If enforcing GAC on a particular global constraint
is very expensive, we may first test whether it is necessary or not to 
launch the propagation algorithm (i.e., whether 
the constraint is already GAC).  On a more academic level,
this question is also commonly asked to
compare different levels of local consistency. 
\begin{quote}
\gacisit(${\cal C}$)\\
\textbf{Instance.} A constraint $C$ of type $\cal C$,  a domain $D$ on $var(C)$\\
\textbf{Question.} Does $\gacsup[C,D,X,v]$ answer ``yes'' for each
  variable $X\in var(C)$ and each value $v\in D(X)$?   
\end{quote}

The third question can be used to
decide if we do not need to backtrack at a given node in the search
tree. Note that $D'\subseteq D$ stands for: $\forall X_i\in var(C),
D'(X_i)\subseteq D(X_i)$. 
\begin{quote}
\gacwo(${\cal C}$)\\
\textbf{Instance.} A constraint $C$ of type $\cal C$,  a domain $D$ on $var(C)$\\
\textbf{Question.} Is there any non empty $D'\subseteq D$ on which
  $\gacisit[C,D']$ answers ``yes''?   
\end{quote}

An algorithm like GAC-Schema \cite{Bessiere-Regin97}
removes values from the initial domain of variables till we
have the (unique) {\em maximal} generalised arc consistent subdomain.
That is, the  subdomain that is GAC and any larger
subdomain is not GAC. 
% or are not included in the initial domains.
The following question characterises this ``maximality'' problem: 
\begin{quote}
\gacmax(${\cal C}$)\\
\textbf{Instance.} A constraint $C$ of type $\cal C$,  a domain $D_0$ on $var(C)$, and a subdomain $D\subseteq D_0$\\
\textbf{Question.} Is it the case that $\gacisit[C,D]$ answers ``yes''
  and there does not exist any domain $D'$,  $D\subset D'\subseteq
  D_0$,  on which $\gacisit[C,D']$ answers ``yes''?
\end{quote}

We finally consider the problem of returning the
domain that a GAC algorithm computes. This is not
a decision problem as it computes something other than
``yes'' or ``no''. 
\begin{quote}
\gacdom(${\cal C}$)\\
\textbf{Instance.} A constraint $C$ of type $\cal C$,  a domain $D_0$ on $var(C)$\\
\textbf{Output.} The domain $D$ such that $\gacmax[C,D_0,D]$ answers ``yes''
\end{quote}

The next subsection shows the intractability of two of the above questions.

\subsection{Intractability of GAC reasoning}\label{sec:intract}

We consider two representative decision problems
at the heart of reasoning with global
constraints. We will show later that their intractability implies 
intractability of the three others. The first is
{\sc GACSupport}, the problem of deciding if a value
for a variable has support on a constraint. In
general, this is NP-complete to decide.

\begin{obs}
{\sc GACSupport} is NP-complete.
\end{obs}
\proof
Clearly it is in NP as a support is a polynomial witness
which can be checked (by our definition of constraint) in polynomial time.
To show completeness, we transform the satisfiability
of the Boolean formula $\varphi$ into the problem of determining
if a particular value has support.
We simply construct the global constraint $C$ involving the
variables of $\varphi$ plus an additional new variable $X$, and defined by
$f_C = (X \rightarrow \varphi)$. If $X=true$ has support
then $\varphi$ is satisfiable.
\qed

The second decision problem we consider is {\sc IsItGAC}.
Given a constraint and domains for its variables, this is the
problem of deciding if these domains are GAC.  This is again a
NP-complete problem. 

\begin{obs}
\textsc{IsItGAC} is NP-complete.
\end{obs}

\proof
Clearly it is again in NP as a support for each value is a polynomial witness
which can be checked in polynomial time since there are $nd$ values
involved where $n$ is the number of variables and $d$ the size of the
largest domain. To show completeness, 
we transform \textsc{3Col}, the problem of deciding whether a graph is
3-colorable into the problem of deciding if a particular 
domain is GAC for a given constraint. 
We introduce a variable for each vertex with
domain $\{r,g,b\}$.
We then define a global constraint as follows.
For each pair $(x_i,x_j)$ of vertices with an edge between in the graph, we permit
pairs of values that are different (i.e., the set
$\{(r,g),$ $(r,b),$ $(g,r),$ $(g,b),$ $(b,r),$ $(b,g)\}$).
For each pair $(x_i,x_j)$ of vertices with no edge between
in the graph, we permit
any pair of values  (i.e., the set
$\{r,g,b\}\times\{r,g,b\}$).
Since values are completely interchangeable, 
$r$, $g$ and $b$ are GAC for a variable iff the graph is 3-colorable. 
Hence, $\{r, g, b\}$ is a GAC domain
for each variable iff the graph is  3-colorable.
\qed

% \begin{corollary}
% Enforcing GAC is NP-hard.
% \end{corollary}
% \proof
% Enforcing GAC directly answers both the \textsc{GACSupport} and \textsc{MaxGAC}
% problems.
% \qed

% % In \cite{besetalAAAI04}, it has been shown that there exists
% % constraint types on which \gacsup\ is NP-complete and \gacmax\ is
% % D$^P$-complete.  In the following, we describe more
% % accurately the relations between the complexity of the different
% % problems we defined. 

% In \cite{besetalAAAI04}, we studied \gacsup\ and \gacmax
% and characterised their complexity in general. 

% \begin{theorem}[\cite{besetalAAAI04}]
% When there is no restriction on the constraint type, \gacsup\ is
% NP-complete. 
% \end{theorem}

% \begin{theorem}[\cite{besetalAAAI04}]
% When there is no restriction on the constraint type, \gacmax\ is
% $D^P$-complete. 
% \end{theorem}

We have proven that two of the questions related to
generalised arc consistency are intractable in general. 
In the following, we  see that there are dependencies between the
intractability  of the five questions. This permits us to deduce that
all five questions are in fact intractable in general. 

\subsection{Intractability relationships}
\label{sec:dependencies}

The five problems defined in Section \ref{sec:questions} are not
independent. Knowledge about intractability of one of them can give
information on intractability of others. We identify here
the dependencies between intractability of the different questions. 

\begin{lemma}\label{lemma:gacsup_gacwo}
$\gacsup(\cal C)$ is NP-hard iff
$\gacwo(\cal C)$ is NP-hard. 
\end{lemma}

\proof
($\Rightarrow$) 
\gacsup$(\cal C)$ can be transformed in 
\gacwo$(\cal C)$: Given $C\in \cal C$,
\gacsup$[C,D,X,v]$ is solved by calling
\gacwo$[C,D|_{D(X)=\{v\}}]$.

($\Leftarrow$)
\gacwo$[C,D]$ can be reduced to
\gacsup\ by calling \gacsup$[C,D,X,v]$ for each
value $v$ in $D(X)$ for one of the  $X$ in $var(C)$. GAC leads to a
wipe out iff  none of these values has a support. \qed

\begin{lemma}
$\gacsup(\cal C)$ is NP-hard iff
$\gacdom(\cal C)$ is NP-hard. 
\end{lemma}

\proof
($\Rightarrow$)
\gacsup$(\cal C)$ can be reduced to
\gacdom$(\cal C)$ since \gacsup$[C,D,X,v]$
answers ``yes'' iff \gacdom$[C,D|_{D(X)=\{v\}}]$ doesn't
return an empty domain.  

($\Leftarrow$)
\gacdom$[C,D]$ can be reduced to \gacsup\ by performing a
polynomial number of calls to \gacsup$[C,D,X,v]$, one for each $v\in
D(X)$, $X\in var(C)$. When the answer is ``no'' the value $v$ is
removed from $D(X)$, otherwise it is kept. The domain obtained at the
end of this process represents the output of \gacdom.   
\qed

\begin{corollary}\label{corol:gacwo_gacdom}
\gacwo$(\cal C)$ is NP-hard iff
\gacdom$(\cal C)$ is NP-hard. 
\end{corollary}

\begin{lemma}
If \gacmax$(\cal C)$ is NP-hard
% \footnote{$D^P$-complete problems are NP-hard. } 
then  \gacsup$(\cal C)$ is NP-hard. 
\end{lemma}

\proof
\gacmax$[C,D_0,D]$ can be reduced to \gacsup. We  perform a
polynomial number of calls to  \gacsup$[C,D_0,X,v]$, one for each $v\in
D_0(X)$, $X\in var(C)$. When the answer is ``yes'' the value $v$ is
added to a (initially empty) set $D'(X)$. \gacmax\ answers ``yes'' if
and only if the domain $D'$ obtained at the
end of the process is equal to $D$.  \qed

\begin{lemma}\label{lemma:gacisit_gacmax}
If \gacisit$(\cal C)$ is NP-hard then 
\gacmax$(\cal C)$ is NP-hard. 
\end{lemma}

\proof
\gacisit$[C,D]$ can easily be transformed into
\gacmax$[C,D,D]$.  \qed

\begin{corollary}
If \gacisit$(\cal C)$ is NP-hard then 
\gacsup$(\cal C)$ is NP-hard. 
\end{corollary}

% \proof
% \gacisit$[C,D]$ can be reduced to \gacsup\ by performing a
% polynomial number of calls to \gacsup$[C,D,X,v]$, one for each $v\in
% D(X)$, $X\in var(C)$. If one of them  answers ``no'' \gacisit\ answers
% ``no'', otherwise it answers ``yes''.   \qed

It is worth noting that whilst intractability of \textsc{IsItGAC} implies
that of \textsc{maxGAC}, this last question may be outside
NP. In fact, \textsc{maxGAC} is $D^P$-complete. 
%The $D^P$ complexity class  contains problems which are the
%conjunction of a problem in 
%NP and one in coNP  \cite{dpcomplete}.
%modifcation by brahim
% modification of brahim's modification by toby: 
%   we pretty much repeat this in the proof!
%
%More precisely, a language $L$ is in the class $D^P$ iff there are two
%languages $L_1 \in NP$ and $L_2 \in coNP$ such that $L= L_1 \cap
%L_2$. %Note that $D^P$ is \emph{not} $NP \cap coNP$. 
% The class $D^P$ is also known as the second level of the Boolean hierarchy,
% BH$\mbox{}_2$. A typical example of a $D^P$-complete decision
% problem is the {\sc Exact Traveling Salesperson Problem}. 
% Given a
% distance matrix and  an integer $B$, {\sc Exact TSP} decides
% if the shortest tour is \emph{equal} to $B$.
% By comparison, \textsc{TSP}
% decides if there is a tour of length $B$ \emph{or less},
% which is NP-complete. 
% We show that {\sc MaxGAC}  is $D^P$-complete.

\begin{theorem}
\textsc{MaxGAC} is $D^P$-complete.
\end{theorem}

\proof
A problem $Q$ is  $D^P$-complete if there exist  $Q_1$ and  $Q_2$ such
that $Q_1$ is NP-complete,  $Q_2$ is coNP-complete, and $Q$ answers
``yes'' iff $Q_1$ and $Q_2$ answer  ``yes''.
We use \textsc{3Col} and \textsc{Un3Col} as $Q_1$ and $Q_2$.
We suppose without loss of generality that $Q_1$ and $Q_2$ both
involve the same set $X$ of vertices. $E_i$ is the set of edges in
$Q_i$.

We introduce a variable for each vertex with
domain $\{r_1,g_1,b_1,r_2,g_2,b_2\}$.
We then define a global constraint as follows.
For each pair $(x_i,x_j)$ of vertices with an edge between
in both $Q_1$ and $Q_2$, we permit
pairs of values that are different but have the same
subscript (i.e., the set
$\{(r_1,g_1),$ $(r_1,b_1),$ $(g_1,r_1),$ $(g_1,b_1),$ $(b_1,r_1),$ $(b_1,g_1),$
$(r_2,g_2),$ $(r_2,b_2),$ $(g_2,r_2),$ $(g_2,b_2),$ $(b_2,r_2),$ $(b_2,g_2)\}$).
For each pair $(x_i,x_j)$ of vertices with an edge between
in  $Q_1$ and not in $Q_2$, we permit
pairs of values that are different for the subscript
$1$, and any combination for subscript $2$ (i.e., the set
$\{(r_1,g_1),$ $(r_1,b_1),$ $(g_1,r_1),$ $(g_1,b_1),$ $(b_1,r_1),$
$(b_1,g_1)\}$ $\cup$ $\{r_2,g_2,b_2\}\times\{r_2,g_2,b_2\}$).
Similarly, for each pair $(x_i,x_j)$ of vertices with an edge between
in  $Q_2$ and not in $Q_1$, we permit
pairs of values that are different for the subscript
$2$, and any combination for subscript $1$.
Finally, for each pair $(x_i,x_j)$ of vertices with no edge between
in $Q_1$ or in $Q_2$, we permit
any pairs of values with the same subscript (i.e., the set
$\{r_1,g_1,b_1\}\times\{r_1,g_1,b_1\} \cup \{r_2,g_2,b_2\}\times\{r_2,g_2,b_2\}$).
By construction,
$r_i$, $g_i$ and $b_i$ are GAC iff $(X,E_i)$ is 3-colorable.
Hence, $\{r_1, g_1, b_1\}$ is the maximal GAC subdomain
for each variable iff $(X,E_1)$ is 3-colorable,
and $(X,E_2)$ is not 3-colorable.
\qed

\begin{figure*}[tbp]
 \centering
 \includegraphics[width=7cm]{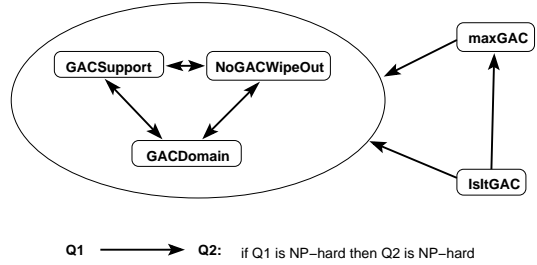}
 \caption{Summary of dependencies between intractable problems\label{figNPH}}
\end{figure*}

\begin{figure*}[htbp]
 \centering
 \includegraphics[width=7cm]{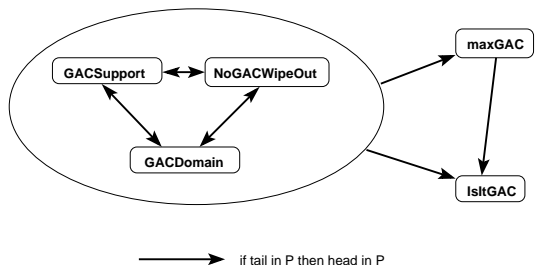}
 \caption{Summary of the dependencies between tractable problems\label{figP}}
\end{figure*}

A summary of the dependencies proved in Lemmas
\ref{lemma:gacsup_gacwo}--\ref{lemma:gacisit_gacmax} and Corollary
\ref{corol:gacwo_gacdom} is given in Fig. \ref{figNPH}.
Note that since each arrow from question $A$ to question $B$ in
Fig. \ref{figNPH} means that $A$ can be rewritten as a polynomial
number of calls to $B$, we immediately have that tractability of $B$
implies tractability of $A$. (See Fig. \ref{figP} for tractability
dependencies of the five questions.)

% *****************************************************
%
% In Sections \ref{sec:counting} and \ref{sec:meta},
% we give a number of constraints for which the
% complexity of GAC was not known. 
% We use the basic 
% tools of computational complexity
% to show their tractability or intractability.

%Assuming $P \neq NP$, r
Reasoning 
with global constraints is thus not tractable in general.
% Note that the dependencies in Fig. \ref{figNPH} still hold in the case
% of bounds  consistency since all
% constructions remain polynomial. They can also be extended to set and multiset
% variables since the occurrence representation introduces a polynomial number
% of variables and bounds consistency on it is equivalent to BC on the set
% or multiset variables \cite{cp03}.
%
Global constraints which are used in practice are therefore
usually part of that special subset for which constraint propagation is
polynomial. For example, GAC on an $n$-ary ${\tt AllDifferent}$
constraint can be enforced in $O(n^{\frac{3}{2}}d)$ time
\cite{regin1}. 
In the rest of this paper, we show how we can
further use the tools of computational complexity in the design
and analysis of {\em specific} global constraints.

% ***[What to do with this?]

% The tools of computational complexity can indeed be used to indicate
% which level of consistency to enforce on a constraint. If achieving a
% given level of consistency is NP-hard, then it is advisable to look
% for a lesser lever of consistency that is polynomial. 

% [etc...]

% We 
% can demonstrate tractability either by showing
% that the constraint can be decomposed into a set of tractable
% constraints without hindering
% propagation, or by exhibiting a polynomial algorithm when the constraint
% cannot be decomposed without hindering constraint propagation. 
% When the constraint is intractable, this
% will be a sufficient condition for the constraint
% not be decomposable (unless P=NP). 

\section{Using Intractability Results}
\label{sec:classify}
The tools of computational complexity can also be used to analyse existing
global constraints for which no polynomial algorithm is known, or can
help us in designing new global constraints for specific purposes. 
To prove that a constraint type $\cal C$ is intractable, 
we generally transform/reduce some known NP-complete/NP-hard
problem to the existence of a satisfying assignment for $\cal C$, i.e.,
the \gacwo$(\cal C)$ problem. Thanks to the dependency
results shown above, we can then deduce intractability of \gacsup\ and
\gacdom. For the more academic questions, \gacisit\ and \gacmax, the
complexity cannot be deduced from our dependencies since they are
'exact' problems (a ``no'' answer brings little information). 
% we cannot easily be sure that a ``no'' answer is correct).  
Finally, we sometimes do not need the full expressive power of a
constraint type to prove its intractability. For example, 
we may use only a fixed value for one of the variables involved in
the constraint. In this case, the constraint is also intractable
if we use its full expressive power. 

%***[Do we introduce bounded domains?]

We can derive several kinds of information about global constraints by
using computational complexity results. For example, 
on existing global constraints for which no polynomial algorithm is
known for a given level of local consistency, proving intractability
tells us that no such algorithm 
exists, % assuming $P \neq NP$, 
and that we should look to enforce a lesser
level of consistency. 
On constraints that decompose into simpler constraints which have polynomial
propagation algorithms, intractability results not only tell us that this
decomposition hinders propagation, but that there cannot exist
any decomposition on which we achieve GAC in polynomial time. % unless $P=NP$. 
We also sometimes want to use an already existing global constraint in a
form more general than its original definition. A proof of
intractability tells us that generalisation makes the constraint
impossible to propagate in polynomial time. %  assuming $P\neq NP$.  
%Finally, tools of computational complexity can be used to answer other
%central questions in constraint programming with the same approach as
%used for
%global constraints. 
%For example, we can answer questions about 
%the tractability of symmetry breaking. 

The remainder of this section gives %a few 
examples of existing and new
global constraints that we analyse with these tools of computational
complexity. 

\subsection{Local consistency}

Computational complexity results can indicate what level of
local consistency to enforce on a constraint. If achieving a given
local consistency on a constraint is NP-hard, then enforcing 
a lower level of consistency may be advisable. 
For example, the number of values
constraint, $\nvalue(X_1,\ldots,X_n,N)$  \cite{pachet1,bcp01}
ensures that $N$ distinct values are used by the $n$ finite domain
variables $X_i$. Note that
$N$ can itself be an integer variable. The $\alldiff$
constraint is a special case of the ${\tt NValue}$ constraint in
which $N=n$. The $\nvalue$ constraint is useful  for
reasoning about resources. 
%***[I DON'T SEE COMPLETELY THE EXAMPLE:
For example, if the values are workers
assigned to a particular shift, we may have a $\nvalue$
constraint on the number of %shifts that someone can work.
workers that a set of shifts can involve. 
Whilst there exists an $O(n^{2.5})$ algorithm for enforcing GAC on the 
$\alldiff$ constraint \cite{regin1}, 
enforcing GAC on the $\nvalue$ constraint is intractable in
general. 

\begin{theorem}
Enforcing GAC on a $\nvalue(X_1,\ldots,X_n,N)$  constraint is NP-hard,
and remains so even if $N$ is ground and different to $n$.
\end{theorem}
%***[Bounded domains]

\proof
% (Sketch) 
We use a transformation from 3{\sc Sat} to \gacwo($\nvalue$). 
Given a 3{\sc Sat} problem in $n$ variables (labelled from
1 to $n$) and $m$ clauses, we construct the 
%number of values 
constraint $\nvalue([X_1,\ldots,X_{n+m}],N)$
in which $D(X_i) = \{ i, -i\}$ for all $i \in [1,n]$,
and each $X_i$ for $i>n$ represents one of the $m$ clauses. 
If the $j$th clause is $x \vee \neg y \vee z$ then
$D(X_{n+j}) = \{ x, -y, z\}$. The constructed 
%number of values 
constraint where $D(N)=\{n\}$ has a solution
iff the original 3{\sc Sat} problem has a satisfying
assignment. 
Hence deciding if enforcing GAC on $\nvalue$ does not lead to a domain
wipe out  
%the existence of a non empty arc consistent domain  
is NP-complete, and enforcing GAC is itself NP-hard. 
\qed

If we want to maintain a reasonable cost for propagation, we therefore
probably have to enforce a lower level of consistency. For instance, there
exists a polynomial algorithm for enforcing bound consistency on the
$\nvalue$ constraint \cite{comicsCPAIOR05}. 

%\subsection{Common constraint}
As a second example, let us take the 
$\common$ constraint, 
$\common(N,M,$ $[X_1,\ldots,X_n],[Y_1,\ldots,Y_m])$
introduced in \cite{beldiceanu3}, that 
ensures that 
$N=|\{i \ | \ \exists j,\ X_i=Y_j\}|$ and
$M=|\{j \ | \ \exists i,\ X_i=Y_j\}|$. That is, 
$N$ is the number of variables in the $X_i$ that take
values in the $Y_j$, and $M$ is the number of 
variables in the $Y_j$ that take values in the $X_i$. 
The $\alldiff$ constraint is again a special case of the
$\common$ constraint in which the $Y_j$ enumerate
all the values $j$ in the $X_i$, $Y_j=\{j\}$ and $M=n$. 

\begin{theorem}
Enforcing GAC on $\common(N,M,[X_1,\ldots,X_n],[Y_1,\ldots,Y_m])$ is NP-hard.
\end{theorem}
% ***[Unbounded domains]
% ***BUT we can modify the proof to have bounded domains: add n+1
% extra X_i variables with domain {dum1, dum2}. Add Y_m+1 with domain
% {dum1}. N=n+1 and M=m+1. 
\proof 
In Theorem \ref{theo:among}, it is shown that enforcing GAC on
$\among(N,[X_1,\ldots,X_n],$ $[D_1,\ldots,D_m])$ is NP-hard, where the
constraint holds iff $N=|\{i \ |  \ \exists j,X_i=D_j\}|$. 
Deciding if enforcing GAC on such an $\among$ constraint does not lead
to a domain wipe out  is equivalent to deciding if enforcing GAC on
$\common(N,M,$ $[X_1,\ldots,X_{n}],[D_1,\ldots,D_m])$ 
with $D(M)=\{0,\ldots,n\}$ does not lead to a domain wipe out. As a result,
enforcing GAC on $\common$  is itself NP-hard. 
%
% We again use a transformation from 3{\sc Sat}. 
% Given a 3{\sc Sat} problem in $n$ variables (labelled from
% 1 to $n$) and $m$ clauses, we construct the $\common$
% constraint, $\common(N,m,[X_1,\ldots,X_{n}],[Y_1,\ldots,Y_m])$
% in which $N = \{0,\ldots,n\}$, $X_i = \{ i, -i\}$, 
% and each $Y_j$ represents one of the $m$ clauses. 
% If the $j$th clause is $x \vee \neg y \vee z$ then
% $Y_{j} = \{ x, -y, z\}$. The constructed 
% $\common$ constraint has a solution
% iff the original 3{\sc Sat} problem has a model. 
% Hence deciding if enforcing GAC does not lead to a domain wipe out 
% %the existence of a non empty arc consistent domain  
% is NP-complete, and enforcing GAC is itself NP-hard. 
\qed

\subsection{Decomposing constraints}
Computational complexity results can tell us more than just
what level of local consistency to enforce. It can also
indicate properties that any possible decomposition of a
constraint must possess. 
%%%March29 modif tried and abandoned
% A set $S_C$ of constraints is a \emph{decomposition} of a constraint
% $C$ if the scope $\bigcup_{C\in S_C}var(C)$ of the decomposition
% contains the scope $var(C)$ of the original constraint, and if the set
% of tuples satisfying $C$ is equal to the projection on $var(C)$ of the
% set of tuples satisfying the  constraints in $S_C$.    
We say that a decomposition of a global constraint is \emph{GAC-poly-time}
if we can enforce GAC on the decomposition in time polynomial in the
size of the original constraint and domains. 
%A decomposition is GAC-poly-time
%when, for example, its number of components
%is polynomial, and each component either is a specific constraint that
%has a polynomial GAC algorithm (like $\alldiff$)
%or has a bounded arity. 
The
following lemma tells us when such decomposition hinders
constraint propagation.
\begin{lemma}\label{theo:decomp_hinder}
If enforcing GAC on a constraint $C$ is NP-hard, then there does not
exist any GAC-poly-time decomposition of $C$ that achieves GAC on
$C$. % (assuming P $\neq$ NP).
\end{lemma}
\proof
By definition, enforcing GAC on a GAC-poly-time decomposition is
polynomial. Hence, if GAC on the decomposition was equivalent to
GAC on the original constraint, then P would equal NP. 
\qed

%\subsection{Disjoint}
Consider a constraint that ensures two sequences of variables
are disjoint (i.e. have no value in common). 
For example, two sequences of tasks sharing
the same resource might be required to be disjoint in time. 
The $\disjoint([X_1,\ldots,X_n],$ $[Y_1,\ldots,Y_m])$ 
constraint introduced in \cite{beldiceanu3}
ensures $X_i \neq Y_j$ for any $i$ and $j$. This constraint has a very
simple and natural decomposition into the set of all binary
constraints $X_i\neq Y_j, i\in [1,n],j\in [1..m]$. Unfortunately, 
enforcing AC on this decomposition into binary constraints
does not achieve GAC on the corresponding $\disjoint$ constraint. 
Consider $X_1, Y_1 \in \{1,2\}$,
$X_2, Y_2 \in \{1,3\}$, and
$Y_3 \in \{2,3\}$. The decomposition into binary constraints is
already AC. 
However, enforcing GAC on $\disjoint([X_1,X_2],[Y_1,Y_2,Y_3])$
prunes 3 from $X_2$ and 1 from both $Y_1$ and $Y_2$.

Moreover, we prove here that we cannot expect any
decomposition to achieve GAC on such a constraint. %, unless P=NP. 
%as it NP-hard to do so in general.

\begin{theorem}
GAC on any GAC-poly-time decomposition
of the $\disjoint$ % ([X_1,\ldots,X_n],[Y_1,\ldots,Y_m])$ 
constraint
is strictly weaker than
GAC on the undecomposed constraint. %(assuming P $\neq$ NP).
%Enforcing GAC on $\disjoint([X_1,\ldots,X_n],[Y_1,\ldots,Y_m])$ is NP-hard. 
\end{theorem}
%***[Bounded domains]

\proof 
We show that enforcing GAC on a $\disjoint$ constraint
is NP-hard, and then appeal to Lemma \ref{theo:decomp_hinder}.
We reduce 3\textsc{Sat} to $\gacwo(\disjoint)$. 
Consider a formula $\varphi$ with $n$ variables
and $m$ clauses. We let $X_i \in \{ i, -i\}$ and
$Y_j \in \{ x,-y,z\}$ where the $j$th
clause in $\varphi$ is $x \vee \neg y \vee z$. If $\varphi$
has a model then the $\disjoint$ constraint
has a satisfying assignment in which the $X_i$ take 
the literals false in this model and the $Y_j$ take the literal
satisfying the $j$th clause. Hence, deciding if enforcing GAC does not lead to a
domain wipe out on $\disjoint$ is NP-complete, and enforcing GAC is itself NP-hard. 
\qed

As another example, Sadler and Gervet
introduce the $\atmostone$
constraint \cite{sadler1}.
This ensures that $n$ set variables of a fixed cardinality $c$
intersect in at most one value.
To fit this within the theoretical framework presented
in this paper, we consider the characteristic
function representation for each set variable
(i.e. a vector of 0/1 decision variables).
Enforcing GAC on such a representation is equivalent
to enforcing bounds consistency on the upper and lower
bounds of the set variables \cite{cp03}.
The $\atmostone$ constraint
can be decomposed into pairwise intersection and
cardinality constraints.
That is, it can be decomposed
into $|X_i \cap X_j| \leq 1$ for $i<j$
and $|X_i|=c$ for all $i$.
On the characteristic function representation,
this is $\sum_k X_{ik} \cdot X_{jk} \leq 1$ and
$\sum_k X_{ik}=c$, which are both GAC-poly-time.
Such decomposition hinders
constraint propagation.
% Besides, we can claim that there is no bounded
% arity decomposition of ${\tt Atmost1}$ that preserves constraint
% propagation (assuming P $\neq$ NP).

\begin{theorem}
GAC on any GAC-poly-time decomposition
of the $\atmostone$ constraint
is strictly weaker than
GAC on the undecomposed constraint. %(assuming P $\neq$ NP).
\end{theorem}
%***[Bounded domains]

\proof
We show that enforcing GAC on an $\atmostone$ constraint
is NP-hard, and appeal to Lemma \ref{theo:decomp_hinder}.
To show that enforcing GAC on the $\atmostone$ constraint
is NP-hard, we consider the case when the cardinality $c=2$.
For $c>2$, we can
use a similar construction as in the
$c=2$ reduction but add $c-2$ distinct values
to each set.
The proof uses a reduction from \textsc{3Sat}.
For each clause $\sigma$,
we introduce a set variable, $X_{\sigma}$.
Suppose $\sigma = x_i \vee \neg x_j \vee x_k$,
then $X_{\sigma}$ has the
domain $\{m_{\sigma}\} \subseteq X_{\sigma} \subseteq
\{ m_{\sigma}, i_{\sigma}, \neg j_{\sigma}, k_{\sigma}\}$.
If the intersection and cardinality
constraint is satisfied, $X_{\sigma}$ takes
the value
$\{ m_{\sigma}, i_{\sigma}\}$,
$\{ m_{\sigma}, \neg j_{\sigma}\}$,
or $\{ m_{\sigma}, k_{\sigma}\}$.
The first case corresponds to $x_i$ being $true$
(which satisfies $\sigma$),
the second to $\neg x_j$ being $true$,
and the third to $x_k$ being $true$.

We use an additional (at most quadratically many)
set variables to ensure that
contradictory assignments are not made to satisfy
other clauses. Suppose
we satisfy $\sigma$ by assigning $x_i$ to $true$.
That is, $X_{\sigma} = \{ m_{\sigma}, i_{\sigma}\}$.
Consider any other clause, $\tau$
which contains $\neg x_i$.
We construct two set variables, $Y_{\sigma \tau i}$
and $Z_{\sigma \tau i}$
with domains
$\{m_{\sigma}\} \subseteq Y_{\sigma \tau i} \subseteq
\{ m_{\sigma}, i_{\sigma}, \neg i_{\tau}\}$
and
$\{\neg i_{\tau}\} \subseteq Z_{\sigma \tau i} \subseteq
\{ m_{\sigma}, m_{\tau}, \neg i_{\tau}\}$.
Since $X_{\sigma} =\{ m_{\sigma}, i_{\sigma}\}$,
then $Y_{\sigma \tau i} = \{ m_{\sigma}, \neg i_{\tau}\}$
and $Z_{\sigma \tau i} = \{ m_{\tau}, \neg i_{\tau} \}$.
Hence, $X_{\tau} \neq \{ m_{\tau}, \neg i_{\tau}\}$.
That is, $\tau$ cannot be satisfied by $\neg x_i$ being
assigned $true$. Some other literal in $\tau$
has to satisfy the clause.

The constructed set variables thus have
a solution which satisfies the intersection
and cardinality constraints iff
the original \textsc{3Sat} problem 
is satisfiable.
Hence deciding if enforcing GAC on $\atmostone$ does not lead to a domain wipe out 
is NP-complete, and enforcing GAC is itself NP-hard.
\qed

A similar result can be given for the ${\tt Distinct}$
constraint introduced in 
\cite{sadler1}.
This constraint ensures that $n$ set variables of a fixed cardinality
intersect in at least one value.
Again, a GAC-poly-time decomposition of such a constraint
hinders constraint propagation. % (assuming P $\neq$ NP).

\subsection{Combining constraints}\label{sec:combining}
%***[Other examples?] 

Global constraints specify patterns that
reoccur in many problems. However, there may only be a limited
number of common constraints which repeatedly occur in problems.
One strategy for developing new global constraints is to identify
conjunctions of constraints that often occur together, and
developing constraint propagation algorithm for their combination.
For example, %Regin and Rueher
\cite{regin4} propose a propagation
algorithm for a constraint which combines
together sum and difference constraints.
As a second example, %Carlsson and Beldiceanu
\cite{lexchain} combine together a chain of lexicographic ordering
constraints.
As a third example, %Hnich, Kiziltan and Walsh
\cite{hkwaimath04}
combine together a lexicographic
ordering and two sum constraints.

We can use results from computational complexity
to determine when we should not combine together constraints.
For example, scalar product constraints
occur in many problems
like the balanced incomplete block design, template design
and social golfers problems  \cite{cp2003}.
Often such problems have scalar product
constraints between all pairs of rows in a 2-dimensional array of 
Boolean decision variables. We can 
enforce GAC on a scalar product constraint between two
rows in linear time.
Should we consider combining together all the row scalar product
constraints into one large global constraint?
Such a  $\sprod$ constraint would ensure that 
$\forall i < j \ \sum_k X_{ik} \cdot X_{jk} = p$.
The following result shows that enforcing GAC on 
such a composition of constraints is intractable.

\begin{theorem}
Enforcing GAC on a $\sprod$ constraint is NP-hard,
even when restricted to 0/1 variables.
\end{theorem}
\proof
%(Sketch)
We consider the case when the scalar product $p=1$. For $p>1$, we use
a reduction that adds $p-1$ additional columns
to the array, each column containing variables that must
take the value 1.

%Note that deciding if GAC does not lead to wipe out on this constraint
%is a special case of the problem [AN9] in \cite{garey}, proved to be
%NP-complete in \cite{frayes77}. 
We reduce \textsc{1in3-3SAT} on positive formulae (which is
NP-complete \cite{garey})
to deciding if enforcing GAC does not lead to a domain wipe out on a 
$\sprod$ constraint 
over 0/1 variables. 
Given a \textsc{1in3-3SAT} problem in $n$ variables  and $m$ clauses,
we construct a $\sprod$ constraint with $4m+1$ rows and $3m+n$
columns. 
The first row of the array, where all variables have 0/1 domain, represents
the model which satisfies the \textsc{1in3-3SAT} problem.
There is a column for each occurrence of a literal in a clause. That
is, the  $(3(j-1)+k)$th column represents the $k$th literal in the $j$th
clause. 
This is assigned 1 in the first row iff the corresponding literal is {\em true}.
There is also a column for the negation of each literal. That is, the
$(3m+i)$th  column represents  the negation of the $i$th  literal.
This is assigned 1 in the first row iff the corresponding literal is {\it false}.
%There are also ``dummy'' columns to ensure each pair of rows has the
%required scalar product. 

The remaining rows are divided into two types.
First, there is a row for each clause.
In the $(1+j)$th row, representing the $j$th clause, the columns
$3(j-1)+1,3(j-1)+2, 3(j-1)+3$ corresponding to literals in the clause
have %0/1 
value 1. The other columns %corresponding to literals
%not in the clause %only 
have the value 0.
The scalar product constraint between a row representing
a clause and the row representing the model ensures
that only one of the literals in the clause is {\em true}.
Second, there are rows for each
occurrence of a positive literal to ensure
that the row representing the model does not assign
both a literal and its negation to {\em true}. 
That is, if the $i$th variable of the formula appears as the $k$th
literal in the $j$th clause, then, in the $(1+m+3(j-1)+k)$th row, 
the columns $3(j-1)+k$ and $3m+i$  
have %0/1 
value 1. The other columns   
have the value 0.

The \textsc{1in3-3SAT} problem has a model
iff the constructed array has
a solution.
Hence deciding if enforcing GAC does not lead to a
domain wipe out on $\sprod$ is NP-complete, and enforcing GAC is NP-hard.
\qed

Special cases of the $\sprod$ constraint
are tractable. For instance, if the scalar product is zero
and variables are 0/1 then
the constraint is equivalent to
the pairwise Disjoint constraint on set variables,
which is tractable \cite{cp03}.

\subsection{Generalising constraints}
Another way in which tools of computational complexity can help is
when we generalise existing global constraints. 
We might have a global constraint with a polynomial propagation algorithm,
but want to use it in a more general manner. 
For example, we might want to replace a given constant parameter 
with  a variable or to repeat the same variable several 
times in the scope of the constraint. 

\subsubsection{Constant parameter becoming a variable}
\label{sec:gen:constant}

%\subsection{Extended global cardinality constraint}
The global cardinality constraint,
$\gcc([X_1,\ldots,X_n],[O_1,\ldots,$ $O_m])$,  ensures that 
$O_j =|\{i~|~X_i = j\}$  for all $j$. That is,
the value $j$ occurs $O_j$ times in the variables $X_i$. 
The special case of this constraint where $O_j$ are fixed intervals 
%given under the form $[l_j,u_j]$, 
was presented in  \cite{regin2} together with a polynomial propagation
algorithm enforcing GAC on the $X_i$. 
The $\alldiff$ constraint is a special case of the
$\gcc$ constraint in which $O_j = [0,1]$. 
However, to enforce GAC on the more general form
of the $\gcc$ constraint where the $O_j$ are integer variables is
NP-hard. 

\begin{theorem}[\cite{quimper03}] 
Enforcing GAC on a $\gcc([X_1,\ldots,X_n],$ $[O_1,\ldots,$ $O_m])$
where the $O_j$ are integer variables  is NP-hard. 
\end{theorem}

%***[Check the domains in the proof of Quimper]

A second example is the $\among$ constraint. The 
$\among(N,[X_1,\ldots,X_n],$ $[d_1,\ldots,d_m])$ constraint, 
introduced in CHIP \cite{beldiceanu2}
ensures that $N=|\{i \ /  \ \exists j,X_i=d_j\}|$. That is, 
$N$ variables in $X_i$ take
values in $[d_1,\ldots,d_m]$. The $\among$ constraint
is a generalisation of the $\atmost$ and $\atleast$
constraints. Enforcing GAC is polynomial on the $\among$
constraint. % \cite{}. 
A  generalisation of this constraint
is to let the $d_j$ be integer variables $D_j$ instead of constants. In this
case, enforcing GAC becomes intractable. 

\begin{theorem}\label{theo:among}
Enforcing GAC on $\among(N,[X_1,\ldots,X_n],[D_1,\ldots,D_m])$ is NP-hard.
\end{theorem}
%***[Bounded  domains]

\proof 
We again use a transformation from 3{\sc Sat}. 
Given a 3{\sc Sat} problem in $n$ variables (labelled from
1 to $n$) and $m$ clauses, we construct the $\among$
constraint, $\among(N,[X_1,\ldots,X_{m}],[D_1,\ldots,D_n])$
in which $D(N) = \{m\}$, $D(D_i) = \{ i, -i\}$, and each $X_j$
represents one of the $m$ clauses.  
If the $j$th clause is $x \vee \neg y \vee z$ then
$D(X_{j}) = \{ x, -y, z\}$. The constructed 
${\tt among}$ constraint has a solution
iff the original 3{\sc Sat} problem has a model. 
Hence deciding if enforcing GAC does not lead to a domain wipe out 
%the existence of a non empty arc consistent domain  
is NP-complete, and enforcing GAC is itself NP-hard. 
\qed

\subsubsection{Repeating variables}
\label{sec:gen:repeat}

%\subsection{Global cardinality constraint with repeated variables}
In the constraint
$\gcc([X_1,\ldots,X_n],$ $[O_1,\ldots,O_m])$, the number
of occurrences $O_j$ for a value $j$ is a fixed interval
$[l_j..u_j]$. In addition, we assume
that no variables in the sequence $[X_1,\ldots,X_n]$
are repeated. 
However, there are problems in which we would
like to have a ${\tt gcc}$ constraint 
with the same variable occurring several
times in $[X_1,\ldots,X_n]$, or equivalently, some variables that must
take the same value. 
For example, in shift rostering, we might have
constraints on the number of shifts worked by each
individual, as well as the requirement that 
the same person works consecutive weekends.
This can be modelled with a $\gcc$ with repeated variables. 
Unfortunately, achieving arc consistency (GAC) on $\gcc$ with
repeated variables  is intractable. 

\begin{theorem}
Enforcing GAC on a $\gcc([X_1,\ldots,X_n],$ $[O_1,\ldots,$ $O_m])$
where  variables in $[X_1,\ldots,X_n]$ can be repeated is NP-hard even
if the $O_j$ are fixed intervals. 
\end{theorem}

%***[Bounded  domains]

\proof
We transform \textsc{3Sat} into \gacwo($\gcc$). 
Let $\varphi=\{c_1,\ldots,c_m\}$ be
a 3CNF on the Boolean variables $x_1,\ldots,x_n$. 
We build the constraint 
$\gcc(Y,$ $[O_{-n},\ldots,$ $O_{-1},O_{1},\ldots,O_n])$ where:
\begin{enumerate}
\item $Y=[Y_{c_1},\ldots,Y_{c_m},
Y^{(1)}_{l_1},\ldots,Y^{(m)}_{l_1},
Y^{(1)}_{l_2},\ldots,Y^{(m)}_{l_n})]$, where
$Y^{(1)}_{l_i},\ldots,Y^{(m)}_{l_i}$ are $m$ copies of the same
variable $Y_{l_i}$
with $D(Y_{l_i})=\{i,-i\}$ and 
$D(Y_{c_j})=\{j_1,-j_2,j_3\}$ if 
$c_j=x_{j_1}\vee \neg x_{j_2}\vee x_{j_3}$,
%\item $V=\{x_1,\ldots,x_n,x'_1,\ldots,x'_n\}$, 
\item $O_i=[0,m], \forall i\in [-n,-1]\cup[1,n]$, 
%\item $E=\{(Y^j_{l_i},Y^{j+1}_{l_i}),  i\in 1..n,j\in  1..m-1\}$. 
\end{enumerate}

%Intuitively, the $Y_{l_i}$ prevent the  $Y_{c_i}$ from taking $i$ as
%value a literal which is true whereas $Y_{l_i}$ 
Consider a model of $\varphi$. If $x_{i_k}$ is one of the variables
in clause $c_i$ that make $c_i$ true in the model, assign $Y_{c_i}$ with  ${i_k}$ if
$x_{i_k}$ is true, and $-{i_k}$ otherwise. For every $i$, assign
$Y_{l_i}$ with  $i$ if 
$x_i$ is false and $-i$ otherwise. This assignment  is a solution 
%consistent tuple 
for  $\gcc$.  

Consider now a solution for $\gcc$. %consistent assignment. 
Then $x_i$ set to true iff
$Y_{l_i}=-i$ is a model of $\varphi$. 
% (for any $j$ because of point (4) of the formulation). 
The $m$ occurrences of each $Y_{l_i}$ and the capacities $O_j$ in the
$\gcc$ ensure that none of the $Y_{c_k}$ can take $-i$ if
$Y_{l_i}=-i$  or $i$ if $Y_{l_i}=i$. 

The constructed ${\gcc}$ constraint with repeated variables  has a
solution iff the original 3{\sc Sat} problem has a model. 
Hence deciding if enforcing GAC does not lead to a domain wipe out 
%the existence of a non empty arc consistent domain  
is NP-complete, and enforcing GAC is itself NP-hard. 
\qed

We see that computational complexity can tell us when 
we will need to enforce a lesser level of consistency on the
generalisation of an existing global constraint.

\section{Meta-Constraints}\label{sec:meta}

Computational complexity can also be used
to study ``meta-constraints'' that combine together other %more primitive
%(possibly non-global) 
constraints.
We will show that even when the constraints being
combined are tractable to propagate, the meta-constraint
itself might not be tractable to propagate.
For example, the $\card$ constraint \cite{cardinality}
is provided by many constraint toolkits. 
%It permits disjunctive to be represented.
It ensures that $N$ constraints from a given set
are satisfied, where $N$ is an integer decision variable.
The most general form
of the constraint is: {$\card(N,[C_1,\ldots,C_m])$}
where {$C_i$} are themselves constraints (not
necessarily all of the same arity), 
%$N = \sum_{i=1}^m |C_i|$,
and $N = |\{i~|~ C_i\textrm{ is satisfied}\}|$. 
%and $|C_i|$ is 1 if $C_i$ is satisfied and zero otherwise.
The cardinality constraint can be used to implement
conjunction,
($C_1 \wedge C_2$ is equivalent to $\card(2,[C_1,C_2])$),
disjunction,
($C_1 \vee C_2$ is equivalent to $\card(N,[C_1,C_2])$ where $N\geq 1$),
negation,
($\neg C_1$ is equivalent to $\card(0,[C_1])$).
It has had numerous applications in a wide range of
domains including car-sequencing, disjunctive
scheduling, Hamiltonian path and digital signal
processor scheduling \cite{vanhentenryck2}.

%(e.g. $M=max(X,Y)$ is equivalent to
%$M\geq X$, $M\geq Y$ and $card(N,[M=X,M=Y,])$ where
%$N \geq 1$).
%

It is obvious that $\card(N,[C_1,\ldots,C_m])$ is 
tractable if the constraints $C_i$ have bounded arity and do not share
any variable. 
However, only a limited form of consistency
is enforced on a $\card$ constraint (see \cite{lhommeCPAIOR04}), and 
%There are four types of propagation events.
%When the upper bound of $N$
%equals $m$, we post each of the constraints, $C_i$.
%When the lower bound of $N$ equals $0$, we
%post the negation of each of the constraints, $\neg C_i$.
%When a constraint $C_i$ is entailed (i.e. it holds
%now for all the remaining values in its domains),
%we can reduce the upper and lower bounds of $N$ by 1
%and eliminate $C_i$ from the cardinality constraint,
%Finally, when the negation of a constraint $C_i$ is entailed (i.e. it
%does not hold for all the remaining values in its domains),
%we can eliminate $C_i$ from the cardinality constraint.
%Not surprisingly, these simple rules do not prune all possible
%values. Indeed,
it is easy to show why %it is necessary
%to enforce only a restricted level of local consistency
%on the cardinality constraint in general.
this is necessary in general. 

\begin{theorem}
Enforcing GAC on the $\card(N,[C_1,\ldots,C_m])$ constraint is NP-hard,
and remains so even if all the constraints $C_i$ are identical and binary and
no variable is repeated  more than three  times.
\end{theorem}
%***[Bounded domains]

\proof
We use a reduction from the special
case of \textsc{3SAT}
in which at most three clauses contain
a variable or its negation. (This is still NP-complete.) 
Each Boolean variable $x$ is represented by a \textsc{CSP}
variable $X$ with domain $\{0,1\}$.
Each clause $\sigma$ is represented by three
\textsc{CSP} variables, $U_\sigma$,
$V_\sigma$ and $W_\sigma$,
and five binary constraints posted on these
variables. The domain of $U_\sigma$ is a strict subset
of $\{8,\ldots,15\}$,
of $V_\sigma$ is a strict subset
of $\{16,\ldots,23\}$ and of
$W_\sigma$ is a strict subset
of $\{24,\ldots,31\}$.
The domain values serve two purposes. First,
the bottom three bits indicate the truth values taken by
the variables that satisfy the clause.
We therefore have to delete one value from each domain.
This is the assignment of truth values which does not
satisfy the clause. For example, if $\sigma$ is $x \vee \neg y \vee z$ then
the only assignment to $X$, $Y$ and $Z$,
which does not satisfy the clause is 0, 1, 0. We therefore
delete the value 26 from $W_\sigma$ as $26 \mymod 8$ is 2 (or 010 in binary).
Similarly, we delete the value 18 from $V_\sigma$ as $18 \mymod 8$ is 2,
and 10 from $U_\sigma$.
Second, the top two bits of the values of $U_\sigma$,
$V_\sigma$ and $W_\sigma$ point to
one of the three positions in the clause.
We add three binary constraints to
the cardinality constraint: $C(U_\sigma,X)$,
$C(V_\sigma,Y)$ and
$C(W_\sigma,Z)$.

We also need to ensure that
$U_\sigma$, $V_\sigma$ and
$W_\sigma$ take consistent values.
We therefore add two binary constraints:
$C(U_\sigma,V_\sigma)$,
and $C(V_\sigma,W_\sigma)$.
Finally, we define $C(X,Y)$ as follows.
If $Y \in \{0,1\}$, there are three cases.
If $8 \leq X \leq 15$ then $C$ is satisfied iff
$(X \mymod 8) \mydiv 4$ = $Y$ (i.e., the third bit of $X$ agrees
with $Y$).
If $16 \leq X \leq 23$ then $C$ is satisfied iff
$(X \mymod 4) \mydiv 2$ = $Y$ (i.e., the second bit of $X$ agrees
with $Y$).
If $24 \leq X \leq 31$ then $C$ is satisfied iff
$X \mymod 2$ = $Y$ (i.e., the first bit of $X$ agrees
with $Y$).
Otherwise $Y \geq 8$ and $C$ is satisfied iff
$X \mymod 8$ = $Y \mymod 8$.

The constructed cardinality
constraint has a solution iff there is
an assignment to the Boolean variables that satisfies all
of the clauses.
Hence enforcing GAC is NP-hard.
\qed

A more restricted, but nevertheless very useful
form of the cardinality constraint is the
cardinality path constraint \cite{cardinality-path}.
The most general form
of the constraint is: {$\cardpath(N,[X_1,\ldots,X_m],C)$}
where {$C$} is a constraint of arity $k$,
and $N = |\{i\in 1..m-k+1~|~C(X_i,\ldots,X_{i+k-1})\textrm{ is satisfied}\}|$.
This ``slides'' a constraint of
fixed arity down a sequence of variables
and ensures that it holds
$N$ times, where $N$ is itself an integer decision variable.
%This slides $C$ down the sequence
%$X_1$, \ldots, $X_m$ and ensure it holds $N$ times.
%The cardinality path constraint can, for example,
This constraint can
be used to implement the change constraint,
(which counts the number of changes of value in a sequence),
smooth constraint
(which limits the size of changes of value along a sequence),
number of rests constraint
(which counts the number of two day or more rests in a sequence),
and sliding sum constraints.
In \cite{cardinality-path}, a greedy algorithm
is given for %partially 
propagating the cardinality path constraint.
However, even for binary constraints, the algorithm fails to prune
all possible values. In \cite{bessiereTR05036}, 
an algorithm is proposed that
achieves GAC when no variable is repeated in the sequence
$[X_1,\ldots,X_m]$ and $C$ has arity $k$. 
This takes a time which is polynomial in $m$ but
exponential in $k$. If $k$ is bounded (e.g. $k=2$),
this is polynomial. The algorithm uses
dynamic programming technique that slides along the values of the
variables the number
of times $C$ can be satisfied in a tuple involving  the given
value. After two passes of this sliding process, the values from $N$
that never appear in the counters can be
pruned, as well as the values that are not labelled by any value in the
domain of $N$. 
As soon as we allow repetitions of variables in the sequence, it
is not hard to show that enforcing GAC on $\cardpath$ is intractable.
As with $\gcc$, this is another example of constraint that
changes from polynomial to intractable when we allow repeated
variables.

\begin{theorem}
Enforcing GAC on $\cardpath(N,[X_1,\ldots,X_m],C)$ where variables in
the sequence $[X_1,\ldots,X_m]$ can be repeated
is NP-hard even if  $C$ is binary. 
\end{theorem}
%***[Bounded  domains]

\proof
We use a reduction from 3{\sc Col}. 
We assume without loss of generality
that the graph is connected. 
Each node in the graph is represented by a CSP variable. 
The domain of each variable is the set of 
three colours. 
We then construct a sequence of variables
$X_1$, \ldots, $X_m$ such that if there is an
edge $(i,j)$ in the graph then there is at least one
position in the sequence with $X_i$ next to $X_j$. 
To do this, we pick any node at which to start. 
We then pick any edge in the graph not yet in
the sequence and find a path from our starting
node that passes through this edge.
We add this path to our sequence. We repeat until all edges 
are in the sequence. 
Finally, we set $N=m-1$
and $C$ to be the binary not-equals constraint.
The constructed cardinality path
constraint has a solution iff there is 
a proper colouring of the graph. 
Hence deciding if enforcing GAC does not lead to a domain wipe out is
NP-complete, and enforcing GAC
is itself NP-hard. 
\qed

It is less easy to see that enforcing GAC on 
$\cardpath(N,[X_1,\ldots,X_m],C)$ is intractable when
the sequence of variables $[X_1,$ $\ldots,X_m]$ does not contain any
repetition and GAC can be enforced on $C$ in polynomial time. 
% On one side, the width of the underlying constraint graph
% is bounded by the arity of $C$, so that
% ${\tt cardpath}(m,[X_1,\ldots,X_m],C)$ is clearly polynomial. On
% the other side, thanks to the domain of $N$, we can express that only
% a part of the occurrences of $C$ in the sequence have to be satisfied,
% which could be source of intractability, as \textsc{2-Sat} which
% becomes intractable when we constrain the number of satisfied clauses
% (e.g., \textsc{Majority-2-Sat}). 

\begin{theorem}
Enforcing GAC on $\cardpath(N,[X_1,\ldots,X_m],C)$
is NP-hard even when enforcing GAC on $C$ is polynomial and  no variable
is repeated in the sequence. 
\end{theorem}
%***[Bounded  domains (to be checked)]

\proof 
We transform \textsc{Max2sat} into
\gacwo($\cardpath$). \textsc{Max2\-sat} is the problem of
deciding whether there exists an assignment of  $n$ Boolean variables
violating at most $k$ clauses in a \textsc{2sat} formula with $m$
clauses. 
The idea is
to build a sequence of variables, alternating $n$ Boolean variables
with two variables representing one of the binary clauses, and then
again $n$ Boolean variables and so on until all clauses are
represented. The sliding constraint $C$ 
guarantees that in each alternation, the assignment of the $n$ Boolean
variables on the left of the two clause-variables is equal to the
assignment on the right (i.e. the same assignment is used
down the sequence), and that the binary clause sandwiched in the
middle is satisfied by this assignment. To prevent violation of a
clause being confused with a change in the assignment, we need $k+1$ dummy
variables in each alternation. A change in the assignment then
violates $k+1$ times the constraint $C$. ($k$ is the bound of the
\textsc{Max2sat} problem.) So, the whole sequence is composed
of $m$ alternations, each with $k+1$ dummy variables plus $n$
Boolean variables plus two clause-variables, plus some additional
dummy variables at the very end of the sequence to guarantee that the
last clause is checked. The domain of the dummy variables
is $\{n+1\}$, that of Boolean variables is $\{0,1\}$. 
% and that of variables representing a literal in a clause is $\{i\}$
% if $x_i$ appears positively in the clause, $\{-i\}$ otherwise.  
%The two clause-variables in the $j^{th}$
%alternation represent the clause $c_j$ in the formula. 
If $c_j=x_{i_1}\lor \neg x_{i_2}$, the first clause-variable in the $j$th
alternation has domain $\{i_1\}$ and the second
has domain $\{-i_2\}$. 
The constraint
$C$, of arity $2(k+1)+2n+4$ (two alternations), is built to be satisfied
in the three following cases:  if neither its first variable nor its
$(k+1+n+2)$th is a
dummy ($k+1+n+2$ is the length of an alternation); 
if its first variable is a dummy and the two
assignments of $n$  Boolean variables are the same; 
finally if the first variable is not a dummy, the $(k+1+n+2)$th variable is,
and the clause represented by the two variables in positions $n+1$ and $n+2$ is
satisfied by the assignment.  
% More formally, $C(X_1,\ldots,X_{2m+2n+4})$ is satisfied iff 
% $$
% ((X_1=n+1)\land \bigwedge_{i=j}^{n+j}(X_{i}=X_{i+2+m+n})), 
% j\textrm{ is the }1^{st}\textrm{ not dummy variable}
% $$
% $$\lor
% $$
% $$ 
% ((X_{2m+2n+3}=n+1)\land 
% ((X_{|X_{n+1}|} \leftrightarrow (X_{n+1}>0))
% \lor(X_{|X_{n+2}|} \leftrightarrow (X_{n+2}>0))))
% $$
% $$\lor
% $$
% $$
% ((X_1\neq n+1)\land(X_{2m+2n+3}\neq n+1))
% $$
Enforcing GAC on $C$ is clearly polynomial. 
% We now need to build the
% sequence of variables repeating $m$ times the alternation
% $X_1...X_{k+1+n+2}$ where  $X_1...X_{k+1}$ are dummy variables,
% $X_{k+2}...X_{k+1+n}$ 
% are Boolean variables, and $X_{k+1+n+1}$ and $X_{k+1+n+2}$ are
% clause-variables.  At
% the very end, we add a last sequence of  dummy variables (of length 2(k+1)+n+2) so that the
% last clause can be checked. 

There remains to set the domain of $N$ to the interval from the total
number of occurrences of $C$ in the sequence (all $C$
satisfied) to this number less $k$. This ensures that $C$ is
violated at most $k$ times. As 
a change in the assignment to the Boolean variables costs
at least $k+1$ violations, we are guaranteed that the same assignment
'slides down' the sequence. Thus $\cardpath$ has
a satisfying tuple if and only if there exists an assignment of the
Boolean variables of the \textsc{Max2sat} formula that violates at most
$k$ binary clauses. Therefore, deciding if enforcing
GAC does not lead to a domain wipe out is
NP-complete,  and enforcing GAC is itself NP-hard. 
\qed

%It remains open to know whether ${\tt cardpath}$ is tractable when $C$
%has a bounded arity. 

We have seen that $\cardpath$ is tractable when $C$ has a fixed arity, 
and we do not allow repetitions of variables in the sequence. 
However, as soon as we relax either one of these 
restrictions, propagation becomes
NP-hard. We may therefore need to enforce a lesser level of local
consistency such as in \cite{cardinality-path}.

\section{Related Work}\label{sec:related}

Analysis of tractability and
intractability is not new in constraint
programming. Identifying properties under which a constraint satisfaction
problem is tractable has been studied for a long time. For
example, Freuder \cite{freuder82}, Dechter and Pearl \cite{decpea88,decpea89}
or Gottlob et al \cite{gotetalIJCAI99} gave increasingly general
conditions on the structure of the underlying (hyper)graph to obtain
a backtrack-free resolution of a problem.  
van Beek and Dechter
\cite{vanBdecJACM95} and Deville et al \cite{devetalIJCAI97} presented
conditions on the semantics of the individual constraints  that make the problem
tractable.  
Finally, Cohen et al \cite{cohetalCP97} showed that when the
constraints composing a problem are defined as  disjunctions of other
constraints of specified types, then the whole problem is tractable. 
However, these lines of research are concerned with a constraint
satisfaction problem as a whole, and do not say much about individual
particular constraints. 

For constraints of bounded arity, asymptotic
analysis has been extensively used to study
the complexity of constraint propagation both in general and
for constraints with a particular semantics. For example,
the GAC-Schema algorithm of \cite{Bessiere-Regin97}
has an $O(d^n)$ time complexity on constraints
of arity $n$ and domains of size $d$,
whilst the GAC algorithm of  \cite{regin1}
for the $n$-ary ${\tt AllDifferent}$
constraint has $O(n^{\frac{3}{2}}d)$
time complexity. 
% By comparison, 
% we have considered here what happens
% when we let the arity of global constraints
% grow. 
These are upper bounds on the cost of GAC in general or on specific
constraints. By comparison, we have characterised here conditions 
under which no polynomial algorithm for GAC can be designed for a
given constraint type. % (assuming $P\neq NP$). 

For global constraints like the
${\tt Cummulative}$ and
${\tt Cycle}$ constraints, there are very immediate reductions from
the bin packing and Hamiltonian circuit which demonstrate
that reasoning with these constraints is intractable in general. 
It is therefore perhaps not surprising
that there has been little comment in the past about
their intractability. However, as we show here, there
are many other global constraints proposed in the past
like ${\tt NValue}$ and $\atmostone$
where a reduction is less immediate, but the constraint
is intractable nevertheless.

%%%% optimisation constraints
In many constraint problems, the goal is not only to satisfy all
the constraints, but also to minimise (or maximise) an %given
objective
function. %These are constraint optimisation problems.
Constraint propagation can be enhanced in these problems by
cost-based filtering where we also remove values that are proven
sub-optimal. \emph{Optimisation constraints}, that combine a
regular constraint of the problem with a constraint on the maximal
value the objective function can take have been advocated  in
\cite{caseau2}.  GAC on such a combined constraint will not only
prune the values having no support on the regular constraint, but
also the values that do not extend to any satisfying assignment of
the constraint improving the given bound. However, as in the case
of combining constraints (see Section \ref{sec:combining}),
%of combining constraints (see Section \ref{sec:combine}),
such compositions have to be handled with care. The
optimisation version of a constraint for which enforcing GAC is intractable
obviously remains intractable (e.g., \cite{SelmanCP03a}). However,
the optimisation version of a constraint for which GAC is
polynomial either remains tractable (e.g., \cite{FLM02,JCRCW02})
or may become intractable. An example of the latter situation   is
the shortest path constraint, which is the optimisation version of
the path constraint \cite{SelmanCP03b}.
% on which GAC is polynomial.
%Whereas examples of tractable optimisation constraints for which
%GAC is ployomial can be found in \cite{FLM02,JCRCW02}.
%Computational complexity can tell us when the optimization version
%of a constraint for which GAC is polynomial becomes intractable.
%This is the case for the
%shorter path constraint, which is the optimization version of the
%path constraint \cite{SelmanCP03b}.

Beldiceanu has proposed a general framework for describing many
global constraints in terms of graph properties on structured
networks of simple elementary constraints \cite{beldiceanu1}. It
is an interesting open question if we can identify properties or
elementary constraints within this framework which guarantee that
a global constraint is computationally (in)tractable.
%
%CAN ANYONE THINK OF ANYTHING ELSE EVEN REMOTELY
%RELATED? THE MORE RELATED WORK, THE BETTER!
%
%HERE IS A POINTER ON THE PAPER I WROTE WITH PASCAL
%NOT SURE WE SHOULD KEEP IT, BUT IN CASE PASCAL WOULD READ THE PAPER
%IT IS PERHAPS BETTER?
%
Finally, computational complexity can help us 
classify the ``globality'' of constraints 
\cite{besvanhCP03}. Indeed, NP-hardness of enforcing
GAC is a sufficient
condition for a constraint to be \emph{operationally GAC-global}
wrt GAC-poly-time decompositions.

\section{Conclusions}\label{sec:conc}

We have studied the computational complexity
of reasoning with global constraints.
We have considered a number of important
questions related to constraint propagation. For example, 
``Does this value have support?'', or
`Is this problem generalised arc-consistent?''. 
% and ``What are the maximal generalised arc-consistent domains?''. 
We identified dependencies between the tractability and
intractability of these questions for finite domain variables and we
have shown that  these questions are intractable
in general. 
%We have shown that it is NP-complete in general
%to determine if a value has support, and $D^P$-complete
%to decide if a subdomain is the maximal generalised
%arc consistent subdomain.
We have then demonstrated how the same tools
of computational complexity can be used
in the design and analysis of specific global
constraints. 
In particular, we have illustrated how computational
complexity can be used to determine  when a lesser level of local consistency
should be enforced, when decomposing constraints
will reduce propagation, 
when constraints
can be combined tractably and when generalisation leads to
intractability. 
We
%Finally, we used these results to 
showed that a wide range of global constraints, 
both existing and new, 
are intractable. In particular, the $\nvalue$ and $\atmostone$
constraints,  the global cardinality constraint with
repeated variables  and the $\common$ constraint, 
are proven here to be intractable. 
% We must therefore look 
% % either to decompose such constraints or 
% to enforce lesser levels of local consistency
% when the size of such
% constraints grows. 
We have also shown how
the same tools can be used to study meta-constraints like the $\cardpath$
constraint. 
In the future, we plan an extensive study of
the computational complexity of global constraints
beyond finite domain variables (e.g. on set and multiset variables).

\subsubsection*{Aknowledgements}
The second and fourth author are members of
the Knowledge Representation and Reasoning Programme
at National ICT Australia. 
NICTA is funded through the Australian Government's
{\em Backing Australia's  Ability} initiative, in 
part through the Australian Research Council. 
The third author is supported by Science Foundation Ireland. 
We thank Marie-Christine Lagasquie for some advice about
reducibility notions.

\bibliographystyle{plain}
%\bibliography{localbib%,../../../biblio/a-z,../../../biblio/pub,biblio}

\end{document}